\documentclass[10pt,twocolumn,letterpaper]{article}

\usepackage[pagenumbers]{cvpr} 

\usepackage{graphicx}
\usepackage{amsmath}
\usepackage{amssymb}
\usepackage{booktabs}
\usepackage{multirow}
\usepackage{pifont}
\graphicspath{{images/}}
\usepackage[pagebackref,breaklinks,colorlinks]{hyperref}

\usepackage[capitalize]{cleveref}
\crefname{section}{Sec.}{Secs.}
\Crefname{section}{Section}{Sections}
\Crefname{table}{Table}{Tables}
\crefname{table}{Tab.}{Tabs.}

\def\cvprPaperID{10425} 
\def\confName{ICCV}
\def\confYear{2023}

\begin{document}

\title{Towards Better Explanations for Object Detection}

\author{Van Binh Truong\textsuperscript{\rm 1}, 
Truong Thanh Hung Nguyen\textsuperscript{\rm 1,2}, 
Vo Thanh Khang Nguyen\textsuperscript{\rm 1},\\ Quoc Khanh Nguyen\textsuperscript{\rm 1}, Quoc Hung Cao\textsuperscript{\rm 1}\\
\textsuperscript{\rm 1}Quy Nhon AI, FPT Software, \textsuperscript{\rm 2}Friedrich-Alexander-Universität Erlangen-Nürnberg\\
{\tt\small \{binhtv8, hungntt, khangnvt1, khanhnq33, hungcq3\}@fsoft.com.vn}
}

\maketitle

\begin{abstract}
Recent advances in Artificial Intelligence (AI) technology have promoted their use in almost every field. 
The growing complexity of deep neural networks (DNNs) makes it increasingly difficult and important to explain the inner workings and decisions of the network. 
However, most current techniques for explaining DNNs focus mainly on interpreting classification tasks. 
This paper proposes a method to explain the decision for any object detection model called D-CLOSE.
To closely track the model’s behavior, we used multiple levels of segmentation on the image and a process to combine them. 
We performed tests on the MS-COCO dataset with the YOLOX model, which shows that our method outperforms D-RISE and can give a better quality and less noise explanation.
\end{abstract}

\section{Introduction}\label{sec:introduction}
Lately, deep neural networks (DNNs) in object detection for images have become popular because of their superior performance in several domains, such as healthcare~\cite{miotto2018deep, nguyen2023towards} and self-driving cars~\cite{do2018real}. 
However, the lack of transparency in decision-making leads to suspicion among end-users, which could negatively affect the widespread AI applications, especially in areas that require trust from users.
Furthermore, newer regulations like the European General Data Protection Regulation (GDPR)~\cite{regulation2018general} strictly require the transparency of using black-box models. 
Thus, a growing chorus of researchers is calling for eXplainable Artificial Intelligence (XAI) methods.
Today, many XAI methods are proposed mainly for classification problems~\cite{das2020opportunities}.
However, explaining object detectors is a big challenge due to the structural differences between the classification and object detection models. 
Several state-of-the-art XAI methods for object detectors are proposed, such as Surrogate Object Detection Explainer (SODEx)~\cite{sejr2021surrogate} and Detector-Randomized Input Sampling for Explanation of Black-box Models (D-RISE)~\cite{petsiuk2021black}.
Yet, these methods meet problems in giving interpretable explanations, tuning hyperparameters for each object without a feature regions' size information, and degraded performance with large datasets.

Hence, in this paper, our main contributions are as follows:
\begin{enumerate}
\item We proposed a new agnostic XAI method for object detectors, called Detector-Cascading multiple Levels of Segments to Explain (D-CLOSE). 
It can explain any object detector's prediction by giving a saliency map that estimates each pixel's importance in the input image to the model's prediction for each individual object.
\item We evaluated the proposed D-CLOSE on the MS-COCO validation dataset~\cite{lin2014microsoft}. 
Results show that D-CLOSE requires less computation time and provides better performance both in classification and localization than D-RISE, the best XAI method for detector model as for as we know.
\item We proposed quantitative and qualitative evaluations for each object-size group to demonstrate the stability of the methods with large datasets.
\item We analyzed D-CLOSE on four cases of the YOLOX's prediction errors~\cite{ge2021yolox} on the MS-COCO dataset.
\item We further evaluated D-CLOSE with real-world images affected by bad conditions, images containing overlapping objects, and different spectra images to demonstrate the method's applicability.
\end{enumerate}

\section{Related Work}\label{sec:related-work}
\subsection{Object Detection Model}
Object detection is an essential field in computer vision that detects the instances of visual objects of a particular class in digital images.
One-stage and two-stage models are the two main approaches to building an object detector~\cite{zou2019object}.
One-stage models are often based on fixed grids and make predictions directly from the input image, such as YOLOX~\cite{ge2021yolox}, a recent release in the YOLO series~\cite{redmon2016you}.
Primarily, YOLOX uses a decoupled head to avoid the problem of collisions between classification and regression branches that reduce model accuracy.
The two-stage model, Faster-RCNN~\cite{ren2015faster}, proposed a set of regions of interest by selected search or Region Proposal Network. 
The proposed regions are sparse as the potential bounding box candidates can be infinite. 
Then, another fine-tuned network processes these regional proposals to decide the final prediction.

\subsection{Explainable AI}
A series of XAI methods were born and divided into two categories based on visualization of the explanations: Pixel-based saliency and Region-based saliency methods~\cite{hartley2021swag}.

\subsubsection{Pixel-based Saliency Methods}
Pixel-based saliency methods measure each pixel's significance score in the input by backpropagating from the prediction to the desired class, such as Gradient~\cite{simonyan2014visualising}, LRP~\cite{bach2015pixel}, and Deep Taylor~\cite{shrikumar2017learning}, which focus on explaining classification models.
A framework called ``Explain to Fix''~\cite{gudovskiy2018explain}, based on Shapley Additive Explanation (SHAP)~\cite{lundberg2017unified}, was proposed to extend the applicability of these methods to object detection problems.
Later, Contrastive Relevance Propagation~\cite{tsunakawa2019contrastive}, an extension of the LRP method, was proposed to explain the output decisions of the Single-Shot Object Detector~\cite{liu2016ssd} by scoring for object classes and offsets in bounding box locations and generating heatmaps highlighting the inputs that contribute significantly to the output. 
However, pixel-based saliency methods, where pixels are scored separately, are often less interpretable ~\cite{wagner2019interpretable}.

\subsubsection{Region-based Saliency Methods}
Region-based saliency methods usually provide a heatmap or regions in the input image representing the key factors contributing to the model's predictions.
Hence, its explanation is comprehensible to end-users rather than optimizing the accuracy of the explanation on individual pixels~\cite{cooper2021believe}.
The first studies showed that the final activation layers often carry complete feature information, which is the main basis for final predictions~\cite{zhou2016learning,selvaraju2017grad}, so several Class Activation Mapping (CAM)-based methods\cite{zhou2016learning, selvaraju2017grad, chattopadhay2018grad,nguyen2022secam} are proposed to calculate the importance of each feature map in the final activation layer of classification models.
Instead of using only one final convolutional layer, Semantic Input Sampling for Explanation (SISE)~\cite{sudhakar2021ada} uses multiple intermediate convolutional layers to provide better spatial resolution and completeness of the explanation.

Other techniques aim to interpret any model regardless of the model's architecture.
For instance, Local Interpretable Model-agnostic Explanations (LIME)~\cite{ribeiro2016should} generates perturbations from subdivided superpixels from the image, then computes the output values and fits into a single regression model to calculate the weights.
Another method using input noise for sampling is Randomized Input Sampling for Explanation of Black-box Models (RISE)~\cite{petsiuk2018rise}, which generates random perturbation masks and transitions through the model, taking the predicted probabilities for the target class as weights for those masks.
Morphological Fragmental Perturbation Pyramid for Black-Box Model Explanations (MFPP)~\cite{yang2021mfpp} generates a saliency map using multiple levels of superpixels~\cite{kapishnikov2019xrai} to sample perturbations and then combines them.

An advantage of utilizing region-based methods is their explanation for model object detectors. Object detectors' output, classification and localization, need some different techniques to explain them.
Furthermore, most object detectors do not use fully connected (FC) layers but convolutional layers, resulting in the ``receptive field'' of the target output being only part of the input image rather than the entire image as in the classification task~\cite{xia2021receptive}.
To our knowledge, there are currently Surrogate Object Detection Explainer (SODEx)~\cite{sejr2021surrogate} and Detector-Randomized Input Sampling for Explanation of Black-box Models (D-RISE)~\cite{petsiuk2021black} can explain both one-stage and two-stage object detectors.
In detail, SODEx uses Surrogate Binary Classifier to convert object detectors' outputs, then uses LIME to explain.
While SODEx only shows the most important regions based on a segmentation algorithm, D-RISE provides a more intuitive and easier-to-understand explanation as a saliency map.
D-RISE leverages the idea from RISE, replacing the weights for each perturbation sample by calculating the similarity between the proposal vectors and the target vector.
However, D-RISE has a major disadvantage; namely, tuning hyperparameters for each object is difficult because we cannot know the size of the feature regions, which causes great obstacles in model execution and evaluation on a larger dataset.
Additionally, methods, like D-RISE, use the way of creating masks from RISE~\cite{petsiuk2018rise}, only gives the best results with rectangular objects; otherwise, D-RISE's performance degrades because it generates random masks based on the mesh~\cite{yang2021mfpp}.

Hence, we proposed D-CLOSE, a new saliency method that can also interpret any object detector, which is more efficient than D-RISE due to using fewer data samples, saving computational time with superior qualitative and quantitative results.

\section{Proposed method}\label{sec:proposed-methods}
\begin{figure*}
    \centering
    \includegraphics[width=.94\linewidth]{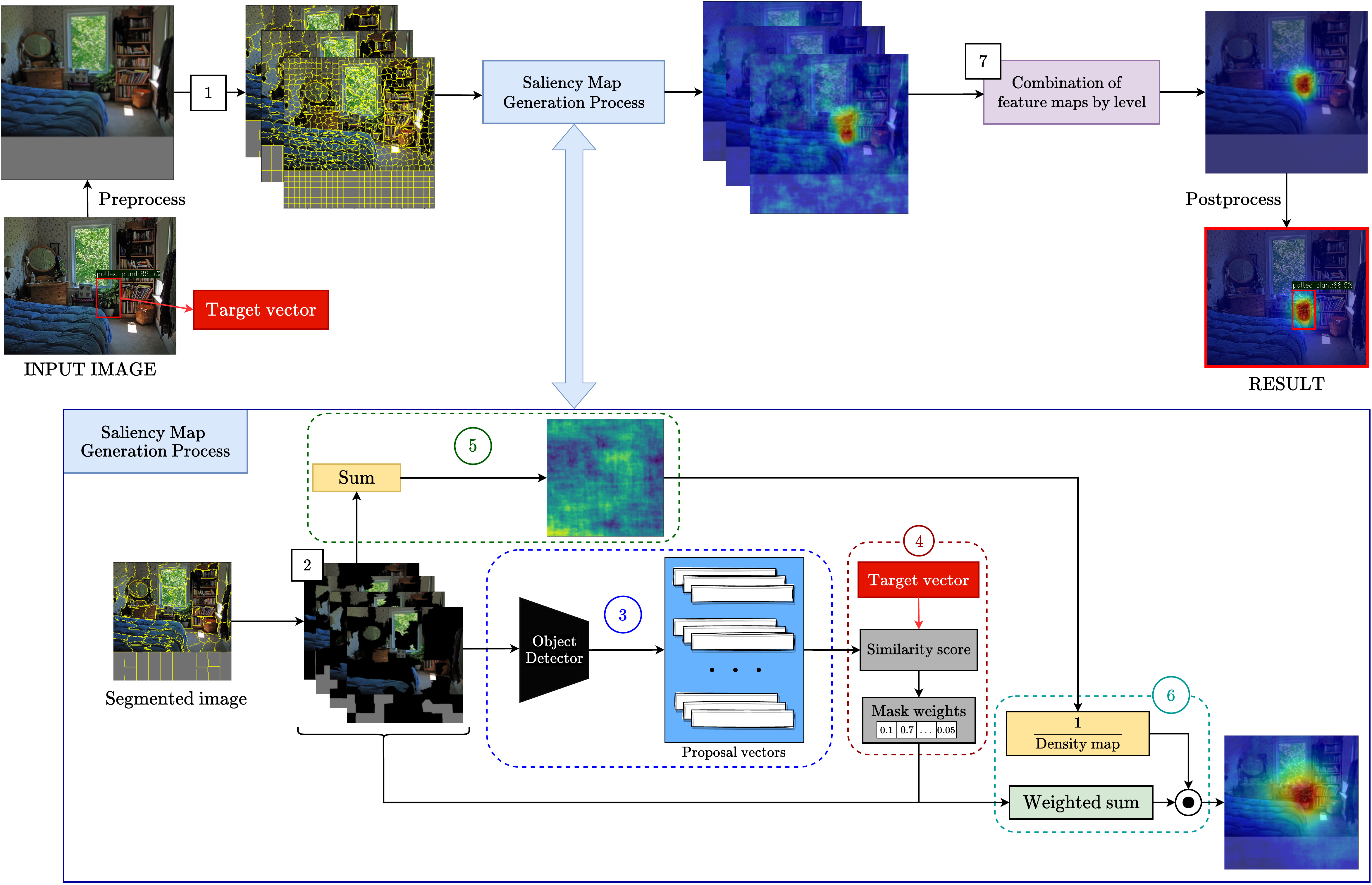}
    \caption{The overall D-CLOSE procedure (upper part) and the detailed saliency map generation process (lower part). Our method builds a standard process for generating saliency maps for segment levels. We work with $L$ different segmentation levels to obtain $L$ feature maps and then aggregate each feature map (shown in Fig.~\ref{fig:sum_feature_map}) to obtain an explanation for the feature. Our method follows seven steps described in Sec.~\ref{subsec_1}.}
    \label{fig: overview_method}
\end{figure*}

We proposed Detector-Cascading multiple Levels of Segments to Explain (D-CLOSE) to generate saliency maps that can explain the decision of both one-stage and two-stage object detectors.
Given input image $I$ of size $h\times w\times 3$, an object detector $D$, our method generates the explanation for detected objects in seven steps ~(Sec.~\ref{subsec_1}). The overall architecture of D-CLOSE is illustrated in Fig.~\ref{fig: overview_method}.

\subsection{Random masks generation}
Images often contain several objects in various sizes and shapes.
We were inspired by mask generation from MFPP~\cite{yang2021mfpp} to generate random masks to explain object detectors.
We inherit the mask generation approach of MFPP as follows:
\begin{itemize}
    \item We use Simple Linear Iterative Clustering (SLIC)~\cite{achanta2010slic}, a quick method to split the image into superpixels with different $L$ levels by changing the number of superpixels [$F_1, F_2,..., F_L$] to segment the image.
    \item We generate $N$ binary masks of size $h\times w$ by setting the segments to $1$ with probability $p$ and $0$ with the remaining segments.
    \item We upsample all masks using bilinear interpolation as this formula $\lfloor (r+1)h \rfloor$$\times$$\lfloor (r+1)w \rfloor$.
    \item We crop masks $h\times w$ with uniformly random indents from $(0,0)$ up to $(\lfloor rh+1 \rfloor$,$\lfloor (rw+1 \rfloor)$.
\end{itemize}

\subsection{Similarity score}
The $d_i$ vector that encodes the predictions of the object detection models is as follows:
\begin{equation}
    d_i = (x^i_1, y^i_1, x^i_2, y^i_2, p_{obj}^i, p^i_1, ..., p^i_C) 
\end{equation}
where:
\begin{itemize}
    \item Detection box (B): coordinates of the predicted objects, $(x^i_1,y^i_1)$ is top-left corner, $(x^i_2,y^i_2)$ is bottom-right corner
    \item Objectness score (O): $p_{obj}^i$ is the probability of predicting a bounding box containing any one object
    \item Detection object's score (C): $(p^i_1,p^i_2,..,p^i_n)$ is the vector representing the correctly predicted scores of the classes in the bounding box.
\end{itemize}

We calculate the correlation between the target vector $d_t$ and the proposal vectors $d_p$, then use it as a weight for the mask, which is forwarded into the model~\cite{petsiuk2021black}. 
We use a similarity score calculation formula from D-RISE:
\begin{equation}
s(d_p, d_t) = IoU(B_p, B_t) \cdot O_p \cdot \frac{C_p C_t}{\| C_p\| \| C_t\|}
\end{equation}


\subsection{Density map}\label{sec:density}
\begin{figure}
    \centering
    \includegraphics[width=.5\textwidth]{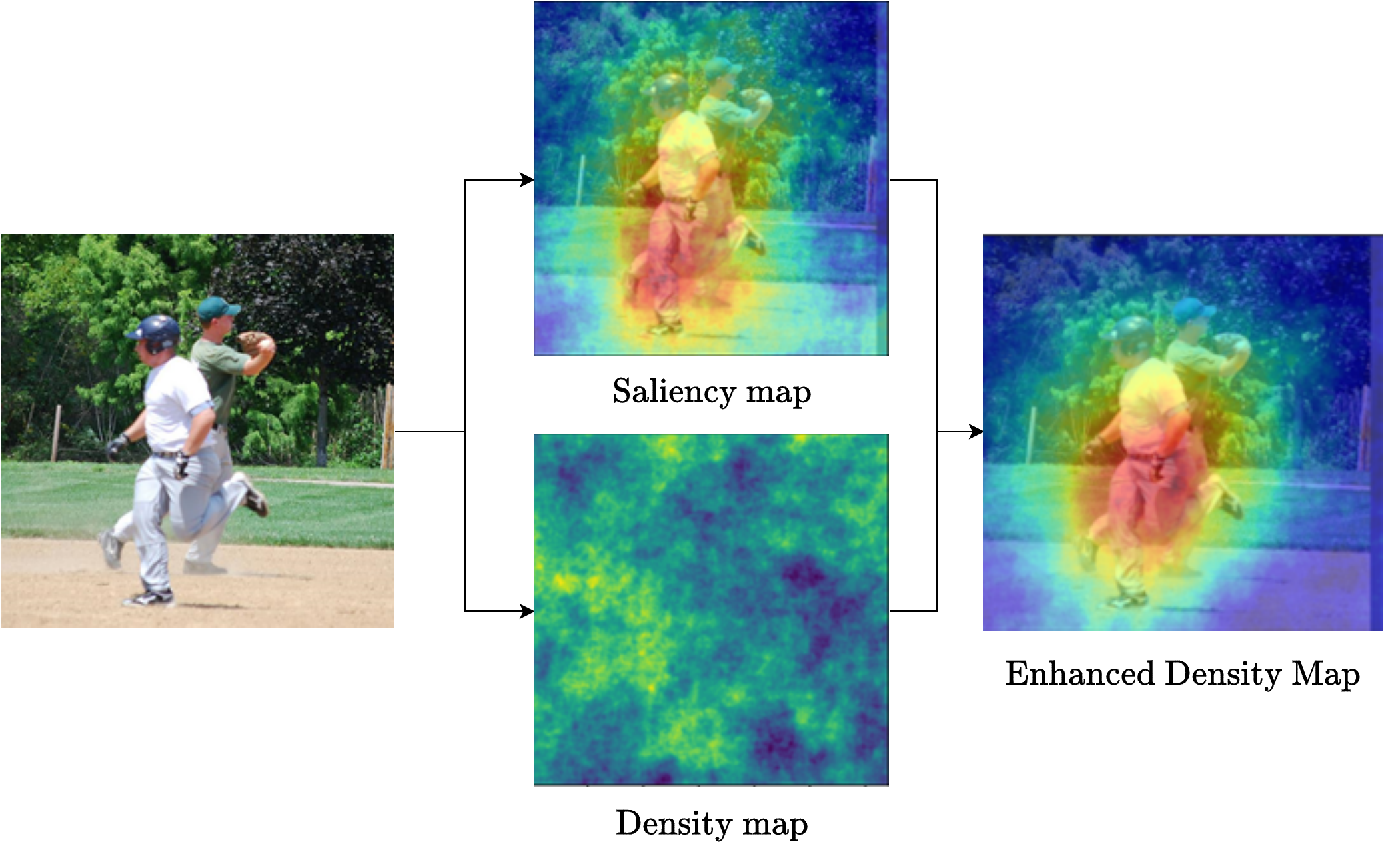}
    \caption{A density map calculates the density distribution of the pixels generated during the masking process. Then, we calculate the average contribution per pixel using a density map to remove noise from the saliency maps.}
    \label{fig:density_map}
\end{figure}

Given an $h$-by-$w$, the mask $M$, $M_{(i,j)}$ is the value of each pixel at position $(i,j)$. Randomly generated masks produce a non-uniform distribution.
Some maybe appear more, and some appear very little, leading to unfair results. 
We aggregate all weighted masks with the output prediction forming a density map $P$ of size $h\times w$ to compute the randomly generated masks' distribution. 
\begin{equation} \label{P}
    P_{(i,j)} = \sum M_{(i,j)} 
\end{equation}
\begin{equation} \label{S}
    S^{'}_{(i,j)} = S^{''}_{(i,j)} \odot \frac{1}{P_{(i,j)}}
\end{equation}
where $S^{''}_{(i,j)}$ and $S^{'}_{(i,j)}$ are the importance scores of each pixel before and after the processing, respectively. 
We have found that normalizing the density map can help produce smoother and less noisy explanations (Fig.~\ref {fig:density_map}). In addition, we also set up experiments and used metric evaluations to strengthen our argument further (Sec.~\ref{sec:ablation}).

\subsection{Fusion feature map}
\begin{figure}
    \fontsize{9pt}{10pt}\selectfont
    \def\svgwidth{\linewidth}
    \centering
\begingroup%
  \makeatletter%
  \providecommand\color[2][]{%
    \errmessage{(Inkscape) Color is used for the text in Inkscape, but the package 'color.sty' is not loaded}%
    \renewcommand\color[2][]{}%
  }%
  \providecommand\transparent[1]{%
    \errmessage{(Inkscape) Transparency is used (non-zero) for the text in Inkscape, but the package 'transparent.sty' is not loaded}%
    \renewcommand\transparent[1]{}%
  }%
  \providecommand\rotatebox[2]{#2}%
  \newcommand*\fsize{\dimexpr\f@size pt\relax}%
  \newcommand*\lineheight[1]{\fontsize{\fsize}{#1\fsize}\selectfont}%
  \ifx\svgwidth\undefined%
    \setlength{\unitlength}{470.99976067bp}%
    \ifx\svgscale\undefined%
      \relax%
    \else%
      \setlength{\unitlength}{\unitlength * \real{\svgscale}}%
    \fi%
  \else%
    \setlength{\unitlength}{\svgwidth}%
  \fi%
  \global\let\svgwidth\undefined%
  \global\let\svgscale\undefined%
  \makeatother%
  \begin{picture}(1,0.74840796)%
    \lineheight{1}%
    \setlength\tabcolsep{0pt}%
    \put(0,0){\includegraphics[width=\unitlength,page=1]{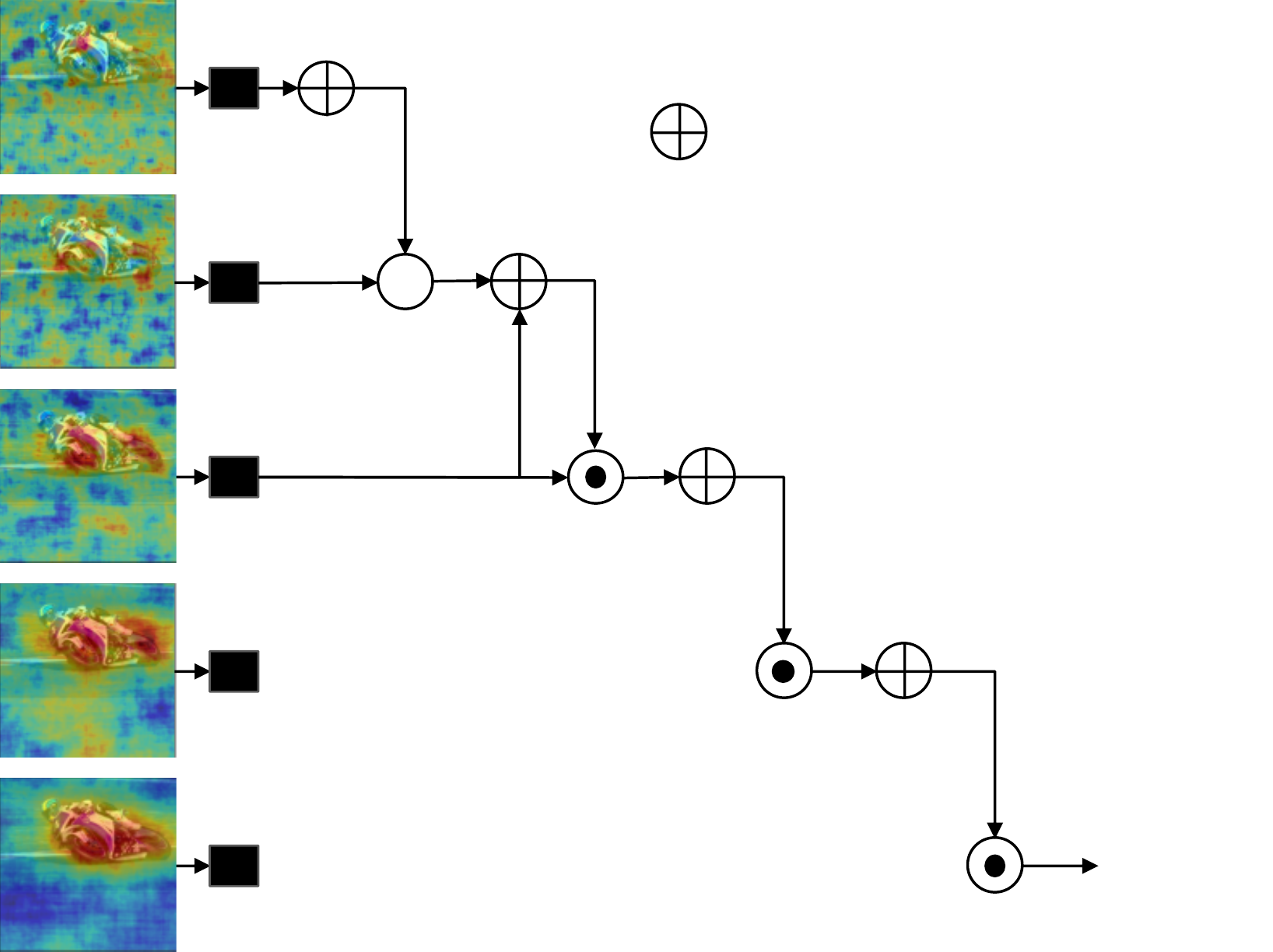}}%
    \put(0.58059901,0.63535058){\makebox(0,0)[lt]{\lineheight{1.25}\smash{\begin{tabular}[t]{l}Addition\end{tabular}}}}%
    \put(0,0){\includegraphics[width=\unitlength,page=2]{sum_all.pdf}}%
    \put(0.58059901,0.57698586){\makebox(0,0)[lt]{\lineheight{1.25}\smash{\begin{tabular}[t]{l}Normalization in range $[0,1]$\end{tabular}}}}%
    \put(0,0){\includegraphics[width=\unitlength,page=3]{sum_all.pdf}}%
    \put(0.58014683,0.5222839){\makebox(0,0)[lt]{\lineheight{1.25}\smash{\begin{tabular}[t]{l}Point-wise multiplication\end{tabular}}}}%
    \put(0,0){\includegraphics[width=\unitlength,page=4]{sum_all.pdf}}%
  \end{picture}%
\endgroup%

    \caption{Our process combines feature maps with multiple levels of segmentation on the image.}
    \label{fig:sum_feature_map}
\end{figure}

After normalizing the density map generated from Sec.~\ref{sec:density}, we obtain $L$ feature maps corresponding to $L$ levels of the superpixel segment.
Each feature map interprets the object's small to large features. 
Intuitively, small features are the most important to identify the object's class, while large features contain the generic and relevant context in which the object is found. 
Our method, inspired by SISE~\cite{sudhakar2021ada}, aggregates feature maps at the semantic level by prioritizing more detailed features and descending to more general features. 
However, SISE forcibly removes noises using threshold parameters with Otsu's algorithm~\cite{otsu1979threshold}, making the explanations sometimes confusing to end-users. 
Additionally, when SISE deals with complex models, selecting the final convolutional layers in the blocks is extremely difficult, especially in object detection problems, because the pooling layers are not at the end of each block. We build a flexible and natural framework (Fig.~\ref{fig:sum_feature_map}), which does not access model internals, does not use threshold parameters, and can construct a wide range of feature maps with areas from detailed to generic regardless of the model architecture.

\subsection{Saliency maps inference}
\label{subsec_1}
We combine all the above operations as a procedure, including random mask generation, similarity score, density map, and fusion feature map, which can infer saliency maps for any object detector.
The procedure is as follows:
\begin{enumerate}
    \item The input image $I$ is divided into $L$ levels of segments. 
    Each segmentation level generated $N$ perturbation masks $M_i^k$, where $1 \leq k \leq L$, $1 \leq i \leq N$.
    
    \item We generate masked images by element-wise masks created with image input ($I \odot M_i^k$).
    
    \item Masked images are forwarded $I \odot M_i^k$ into the object detectors $D$ to get $T$ vectors of prediction:
    \begin{equation}
        d_p = D(I \odot M_i^k) = (d^j_i)^k 
    \end{equation}
    where $1\leq j \leq T, 1\leq i \leq N, 1\leq k \leq L$.
    
    \item We calculate the similarity between target vector $d_t$ to be explained and proposal vectors $d_p$. Then, we take the maximum score for each masked image on each target vector:
    \begin{equation}
        w^k_i = max(s(d_t,(d^j_i)^k))
    \end{equation}
    where $1\leq j \leq T, 1\leq i \leq N, 1\leq k \leq L$.
    
    \item We calculate the density map $P_k$ representing the distribution of randomly generated masks $M^k_i$ in Eq.~\ref{P}, where $1\leq k \leq L$.
    
    \item We compute the average score of each pixel by dividing the weighted sum of masks $M^k_i$ by the density map $P_k$ to obtain the saliency map $S_k$ for the $k^{th}$ segment level.
    \begin{equation}
        S_k = \frac{1}{P_k} \odot \sum_{i=1}^{N}{w^k_iM^k_i}
    \end{equation}
    where $1 \leq k \leq L$.

    \item We combine each semantic level's $S_k$ saliency maps from the features by cascading in blocks, as shown in Fig.~\ref{fig:sum_feature_map}. Mathematically, our process works like a recursive algorithm where $A_k$ is the saliency map obtained after each step. 
    In the last step, the saliency map obtained is $A_{L-1}$:
    \begin{equation}
        A_{k} = 
        \begin{cases}
             (S_k + S_{k+1})S_{k+1}&{k=1}\\
             (A_{k-1}+S_{k+1})S_{k+1}&{2\leq k \leq L-1}\\
        \end{cases}
    \end{equation}
\end{enumerate}

\section{Experiments and results}\label{sec:experiments and results}
\subsection{Datasets and models}
We evaluate our method with a pre-trained YOLOX model~\cite{ge2021yolox}\footnote{\label{1}\url{https://github.com/Megvii-BaseDetection/YOLOX}} on the MS-COCO validation dataset~\cite{lin2014microsoft}. 
In quantitative evaluation, results are obtained with the same set of parameters on the same data set. 
All experiments use Nvidia Tesla T4 and 24GB RAM as the benchmark.

\subsection{Parameters setup}
For D-RISE, we use default parameters proposed in the original paper with $N = 5000$ masks, probability $p=0.5$ and resolution $(h_s, w_s)=(16, 16)$. 

For D-CLOSE, during the masking process, we use a segmentation algorithm that splits the image into $L=5$ levels with the number of segments $[150,$ $300,$ $600,$ $1200,$ $2400]$, respectively. 
For each image, we perform our method with $N=800$ masks per segmentation level, a kernel width $\alpha=0.25$, and resize offset ratio $r=2.2$.

\subsection{Sanity checks}
\begin{figure}[tbh!]
    \centering
    \begin{subfigure}[b]{0.32\linewidth}
        \centering
        \includegraphics[width=\textwidth]{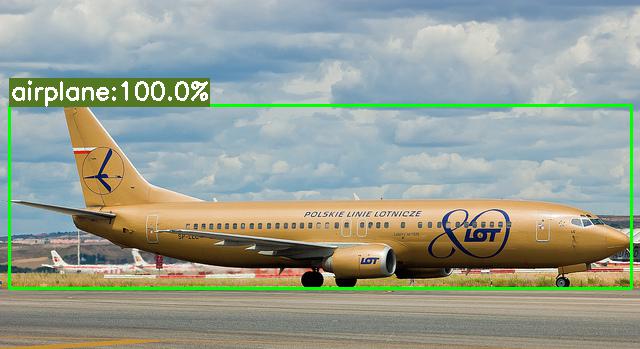}
        \caption{}
        \label{fig: sanity predict}
    \end{subfigure}
    \begin{subfigure}[b]{0.32\linewidth}
        \centering
        \includegraphics[width=\textwidth]{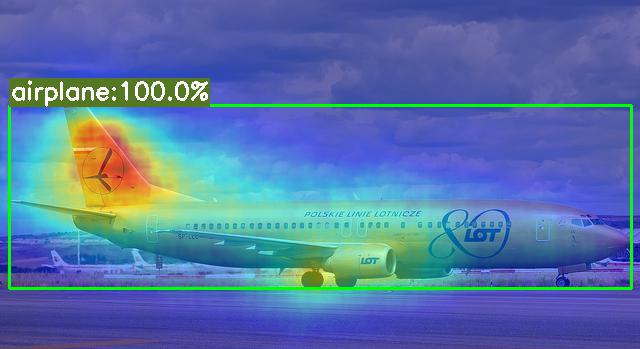}
        \caption{}
        \label{fig: saliency map}
    \end{subfigure}
    \begin{subfigure}[b]{0.32\linewidth}
        \centering
        \includegraphics[width=\textwidth]{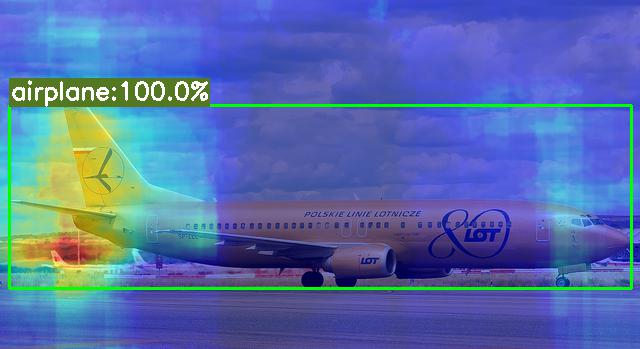}
        \caption{}
        \label{fig: saliency map 2}
    \end{subfigure}
    \caption{(a) Ground truth, (b) Saliency map with pre-trained weight, (c) Saliency map with altered weight.}
    \label{fig:lime_sample}
\end{figure}

A saliency map can explain which features the model considers to predict and why the model gives the correct explanation or not.
Hence, the saliency map reveals the weights that the model learned during training. 
We use sanity checks~\cite{adebayo2018sanity} to check whether D-CLOSE results faithfully reflect the decision-making behavior of the model.
We check whether the saliency map changes by changing the model's weights.
In the experiment, our method's saliency map shows that the model focuses on another region with altered weights, which means that our method can faithfully reflect the model's behavior (Fig.~\ref{fig:lime_sample}).

\subsection{Model errors}
Inspired by~\cite{hoiem2012diagnosing}, we evaluate whether D-CLOSE can generate feature maps that explain the incorrect decisions of the model.
We further extend the test case where the model predicts an object that does not exist in the image.
Also, we analyze the classification error and localization error of the YOLOX model on the MS-COCO dataset.

\begin{figure*}
    \centering
    \includegraphics[width=.95\linewidth]{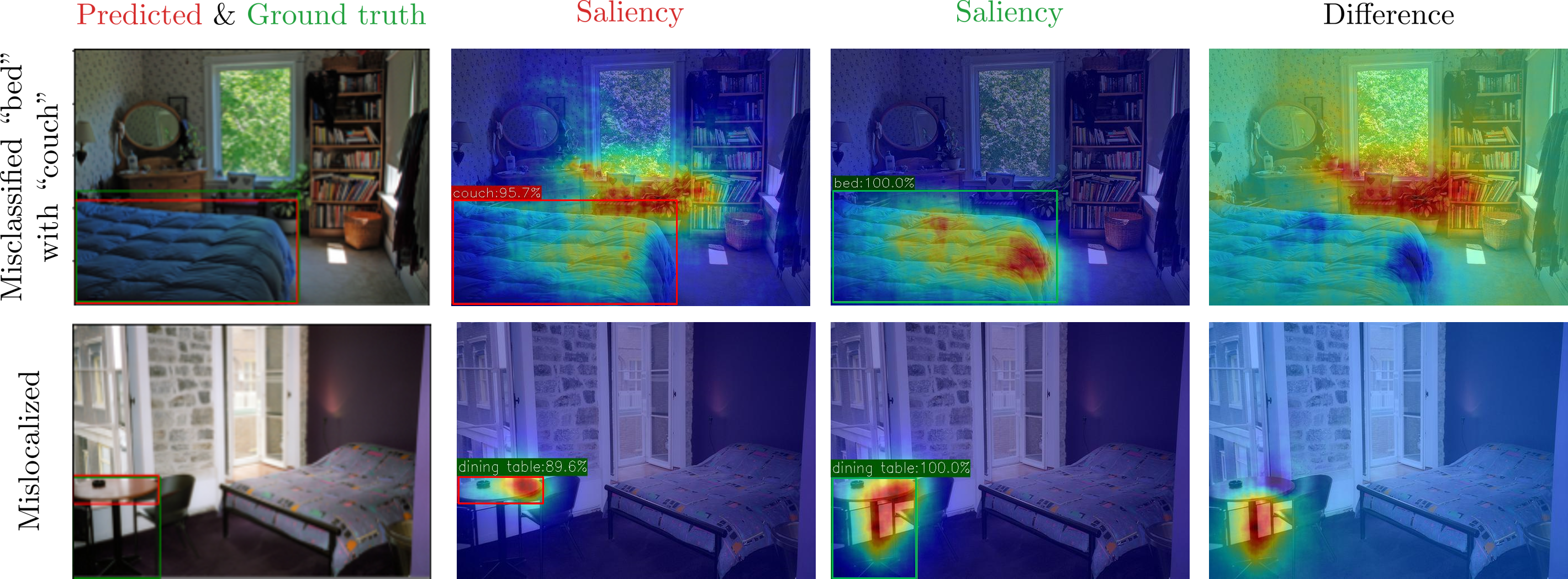}
    \caption{Examples of localization and classification error prediction model cases. The green box is the ground truth, the red box is the model's prediction. In the first row, the model is biased toward the outside context and misclassifies the ``bed'' as the ``couch''. In the second row, the model correctly predicts the ``dining table'', but the model only focuses on the tabletop and ignores table legs leading to poor localization.}
    \label{fig: poor_predict}
\end{figure*}

\begin{figure}[tbh!]
    \begin{subfigure}[b]{0.39\linewidth}
        \centering
        \includegraphics[width=\textwidth]{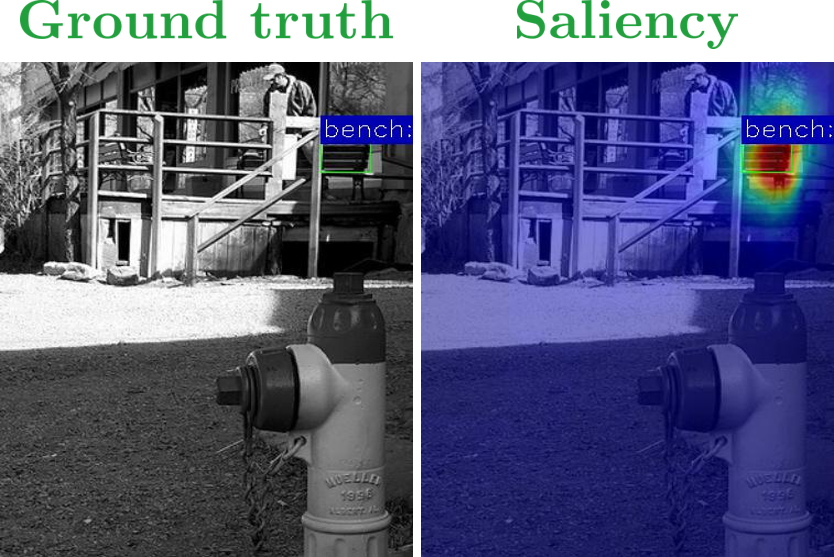}
        \caption{Fail to detect}
        \label{fig:fail_detect}
    \end{subfigure}
    \begin{subfigure}[b]{0.59\linewidth}
        \centering
        \includegraphics[width=\textwidth]{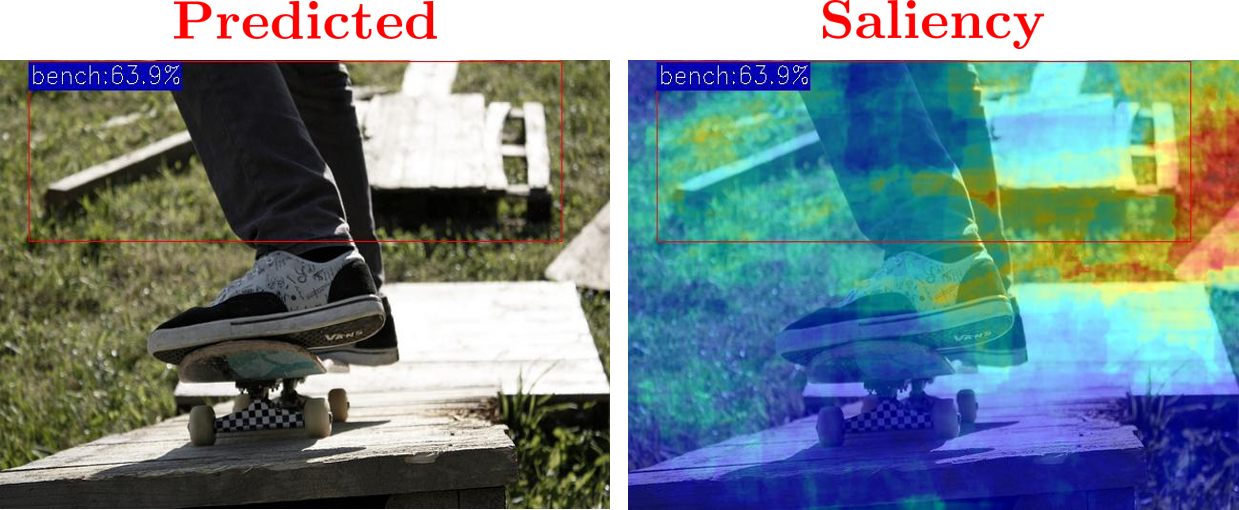}
        \caption{Detect unlabeled object}
        \label{fig:detect_unclassified}
    \end{subfigure}
    \caption{In (a), the model fails to detect the labeled ``bench'', but the explanation shows that the model can still capture the bench's features because possibly those features are not strong enough to influence the model's final decision. In (b), the model detects an object not labeled as ground truth. The explanation shows that the model focuses mostly on the ground (not the wooden planks) to predict the ``bench''.}
\end{figure}

\begin{itemize}
    \item Object is  correctly localized but misclassified (The first row in Fig.~\ref{fig: poor_predict}).
    \item Object is detected with a correct classification but mislocalized (The second row in Fig.~\ref{fig: poor_predict}).
    \item The model fails to detect the object (Fig.~\ref{fig:fail_detect}).
    \item The model detects an object not labeled as ground truth (Fig.~\ref{fig:detect_unclassified}).
\end{itemize}

We compare the two generated explanations for the ground truth and predicted bounding box, then calculate the difference by subtracting the corresponding pixel values between them.
The difference between the two saliency maps can indicate the source of the error.

\subsection{Model in images affected by bad conditions}
Our experiment utilizes D-CLOSE to explain the model's detection of images from surveillance cameras that record images of pedestrians and vehicles in bad conditions such as low light, fog, and night.
Also, based on ~\cite{bayer2021comparison}, we evaluate D-CLOSE with images in different spectra.
Test images are obtained from the WIDER Pedestrian Detection Dataset~\cite{zhang2019widerperson} and Multispectral Object Detection Dataset~\cite{takumi2017multispectral}.

\begin{figure*}
    \centering
    \includegraphics[width=\linewidth]{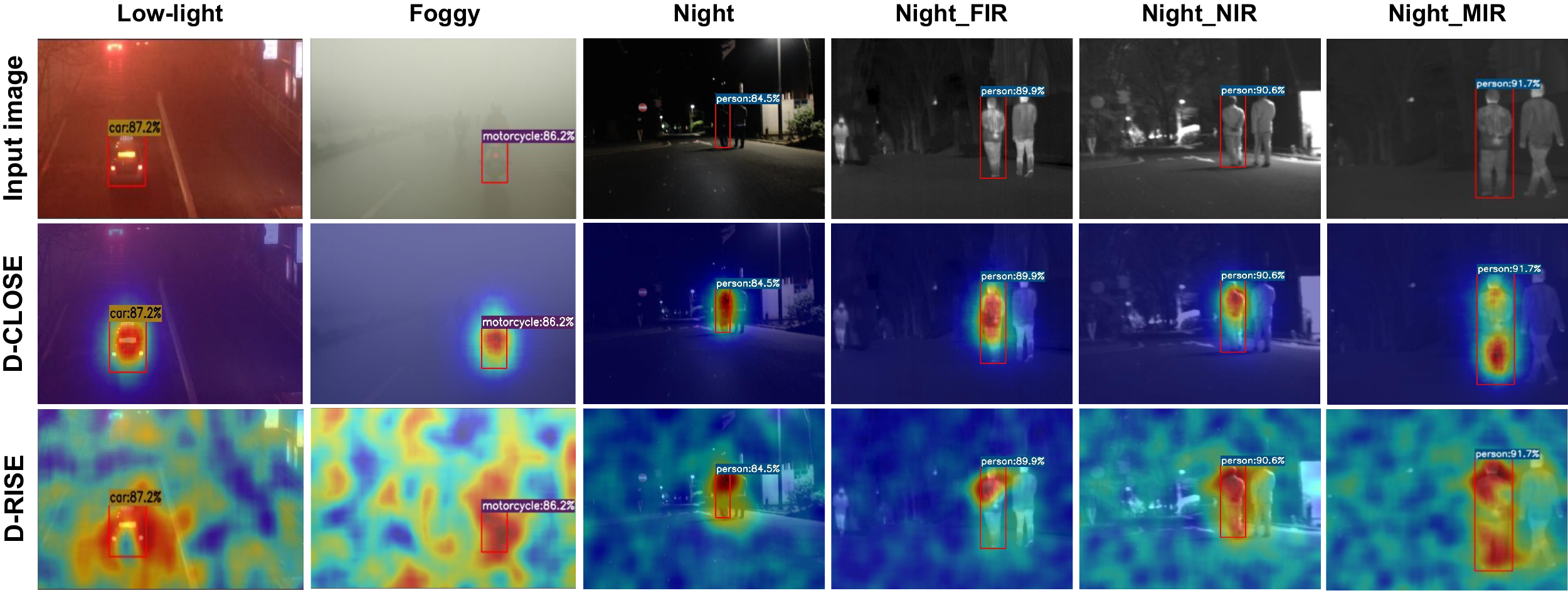}
        \caption{D-CLOSE provides object-focused explanations in images affected by bad conditions: Low-light (Underexposed and affected by red light), Foggy, Night, and in different Night spectra: FIR, NIR, MIR ~\cite{bayer2021comparison}.}
    \label{fig: test_noise}
\end{figure*}

In all cases (Fig.~\ref{fig: test_noise}), D-CLOSE's saliency map is high quality and stable.
While D-RISE produces a more noisy saliency map and is unfocused.
Because our method builds on combining feature maps of multiple segmentation levels, each error map level can be entirely offset by other maps.

\begin{figure}
    \fontsize{8pt}{10pt}\selectfont
    \def\svgwidth{\linewidth}
    \centering   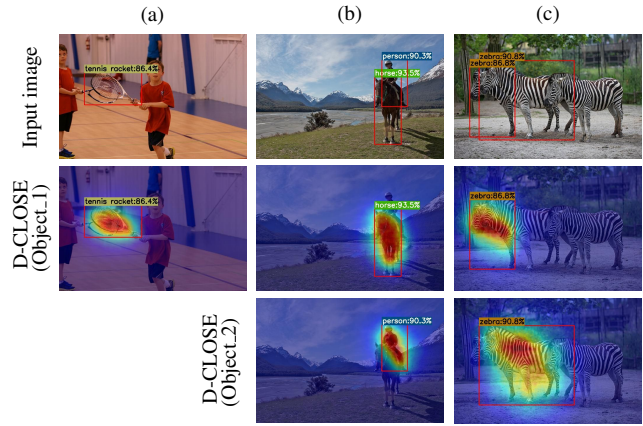
    \caption{Examples of partially and fully overlapping objects. In (a), the bounding box wraps two ``tennis racket'' objects, D-CLOSE indicates that the model mostly focuses on the front ``tennis racket''. In (b) and (c), D-CLOSE generates differentiated saliency maps for overlapping objects.}
    \label{fig: test_overlap}
\end{figure}

\section{Evaluation metrics}\label{sec:evaluation}
\subsection{Qualitative}
This section evaluates the visualizations from D-CLOSE in two approaches: the stability of the method when changing parameters compared to previous methods and the object discrimination ability of our method.
\subsubsection{Stability visualization}
\begin{figure*}
    \centering
    \includegraphics[width=\linewidth]{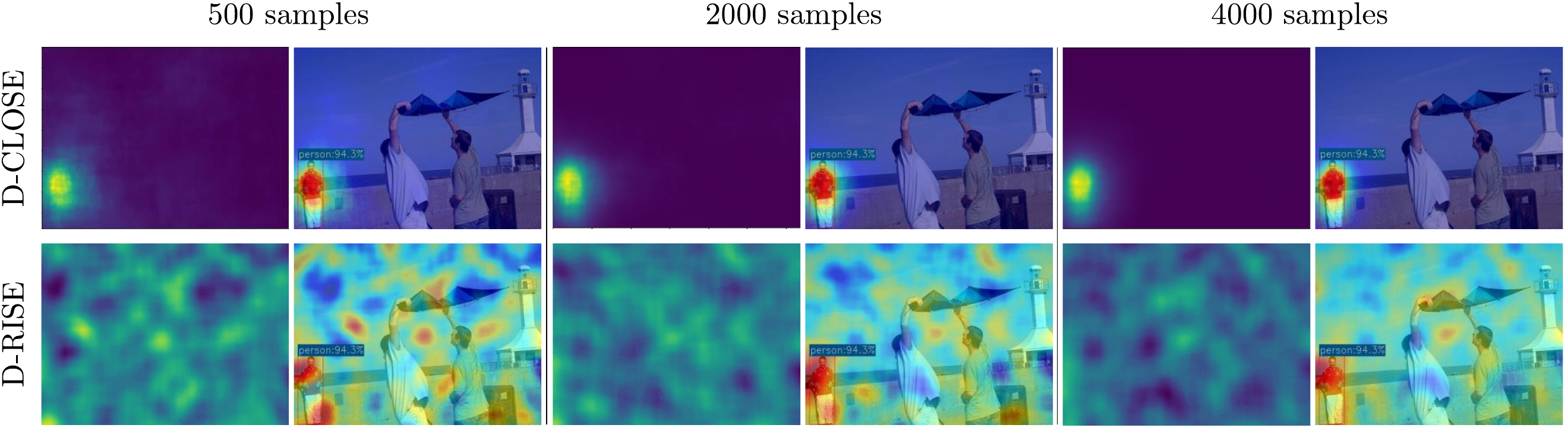}
    \caption{We compare two methods, D-CLOSE and D-RISE, with different numbers of samples of 500, 2000, and 4000 samples, respectively. D-CLOSE produces stable and better-quality saliency maps than D-RISE, even with a small number of samples.}
    \label{fig:compare}
\end{figure*}

Since XAI methods can perform differently with varying number of samples~\cite{nguyen2021evaluation}, we compare D-CLOSE with D-RISE in different amounts of data (Fig.~\ref{fig:compare}) to show that our method produces better results without being influenced by the number of generated samples.
When evaluating both methods on large datasets, we use a fixed set of parameters for all objects in the image.
We found out that D-RISE produced good results only by fine-tuning each parameter to fit each object's geometry; otherwise, D-RISE's saliency map is quite unfocused and provides weak localization in some cases.
While D-CLOSE's explanation is more stable with the number of samples generated, the noise in the saliency map decreases, and more focus is on the important regions in the image.
\subsubsection{Object discrimination visualization}
We conduct experiments to measure whether our proposed D-CLOSE has good object discrimination ability. As shown in Fig.~\ref{fig: test_overlap}, D-CLOSE's explanations focus more on the object's shape inside the bounding box. They often clearly distinguish the boundaries of objects, significantly when multiple objects are overlapped.

\subsection{Quantitative} \label{sec:metric_1}
We apply various metrics to compare our method's plausibility and faithfulness with other methods.

\subsubsection{Plausibility Evaluation}
We use two standard metrics to evaluate XAI's plausibility: \textit{Sparsity}~\cite{gomez2022metrics} and \textit{Energy-based pointing game (EBPG)}~\cite{wang2020score}, based on human-annotated bounding boxes.
In our evaluation, we only consider explanations for detected bounding boxes that best match the ground truths for each class to compute these metrics.

\subsubsection{Faithfulness Evaluation}
Faithfulness evaluation metrics, including \textit{Deletion}, \textit{Insertion}~\cite{petsiuk2018rise} and \textit{Over-all}~\cite{zhang2021group} measure the explanation's completeness and consistency for the model's predictions.
\textit{Deletion} checks whether removing these important pixels severely degrades the model's predictions for that object. For \textit{Insertion}, it measures the increase in probability as more important pixels are included. 
\textit{Over-all} score is the difference between \textit{Insertion} and \textit{Deletion}.

\begin{table}[hbt!]
\centering
\label{tab:2}
\resizebox{\linewidth}{!}{%
\begin{tabular}{ccc|c|c||c|}
\hline
\multicolumn{3}{|c||}{Ablation Settings} & \multirow{2}{*}{Sparsity} & \multirow{2}{*}{EBPG (\%)} & \multirow{2}{*}{Over-all (\%)} \\ 
\cline{1-3}
\multicolumn{1}{|c|}{Superpixel Segment} & \multicolumn{1}{c|}{Density Map} & \multicolumn{1}{c||}{Feature Fusion} &  &  &  \\ \hline
\multicolumn{1}{|c|}{\ding{51}} & \multicolumn{1}{c|}{\ding{55}} & \multicolumn{1}{c||}{\ding{55}} & $3.42 $ & $12.25 $ & $87.31$ \\ 
\multicolumn{1}{|c|}{\ding{51}} & \multicolumn{1}{c|}{\ding{51}} & \multicolumn{1}{c||}{\ding{55}} & $3.93 $ & $13.42 $ & $87.35$ \\
\multicolumn{1}{|c|}{\ding{51}} & \multicolumn{1}{c|}{\ding{55}} & \multicolumn{1}{c||}{\ding{51}} & $22.01$ & $32.52$ & $88.08$\\
\multicolumn{1}{|c|}{\ding{51}} & \multicolumn{1}{c|}{\ding{51}} & \multicolumn{1}{c||}{\ding{51}} & \textbf{25.02} & \textbf{35.45}  & \textbf{88.14} \\\hline
\end{tabular}%
}
\caption{Quantitative evaluation metrics on different ablation settings. The best is shown in bold.}
\label{ablation}
\end{table}

\begin{table*}[hbt!]
\resizebox{\textwidth}{!}{%
\begin{tabular}{@{}ccccccccccccc@{}}
\toprule
& \multicolumn{3}{c}{Small} & \multicolumn{3}{c}{Middle} & \multicolumn{3}{c}{Large} & \multicolumn{3}{c}{Small+Middle+Large} \\
\cmidrule(lr){2-4} \cmidrule(lr){5-7} \cmidrule(lr){8-10} \cmidrule(lr){11-13}
\textbf{Metrics} & Grad-CAM & D-RISE &  D-CLOSE & Grad-CAM & D-RISE &  D-CLOSE & Grad-CAM & D-RISE &  D-CLOSE & Grad-CAM & D-RISE &  D-CLOSE \\
\midrule
\textbf{Sparsity} $\uparrow$        &21.46&4.81&\textbf{28.00}&9.22&2.67&\textbf{12.18}&7.80&2.41&\textbf{8.49} &19.47 & 4.43 & \textbf{25.02}    \\
\midrule
\textbf{EBPG (\%)} $\uparrow$          &11.52&0.06&\textbf{28.34}&59.11&26.89&\textbf{69.50}&81.39&55.57&\textbf{84.22} &26.86 &11.17 &\textbf{35.45}\\
\midrule
\textbf{Del (\%)} $\downarrow$       &3.27&2.27&\textbf{1.21}&13.92&12.28&\textbf{6.53}&25.07&17.43& \textbf{12.57}&5.73 &4.26 &\textbf{2.71}       \\
\midrule
 \textbf{Ins (\%)} $\uparrow$ &71.13&83.27&\textbf{92.35}&68.78&73.06&\textbf{85.74}&62.03&64.05&\textbf{78.92}&70.18 &81.19 &\textbf{90.85} \\
\midrule
 \textbf{Over-all (\%)} $\uparrow$ &67.86&81.00&\textbf{91.14}&54.86&60.78&\textbf{79.21}&36.96&46.62&\textbf{66.35}&64.45 &76.93 &\textbf{88.14} \\
\bottomrule
\end{tabular}
}
\caption{Mean accuracy of quantitative results of all XAI methods evaluated on the whole MS-COCO validation set, further categorized into small, middle, and large groups (as shown in Fig.~\ref{fig:example}). For each metric, the best is shown in bold. The arrows $\uparrow/\downarrow$ indicate higher or lower scores are better.}
\label{tab:quantitative_results}
\end{table*}

\begin{figure}
    \centering
    \includegraphics[width=\linewidth]{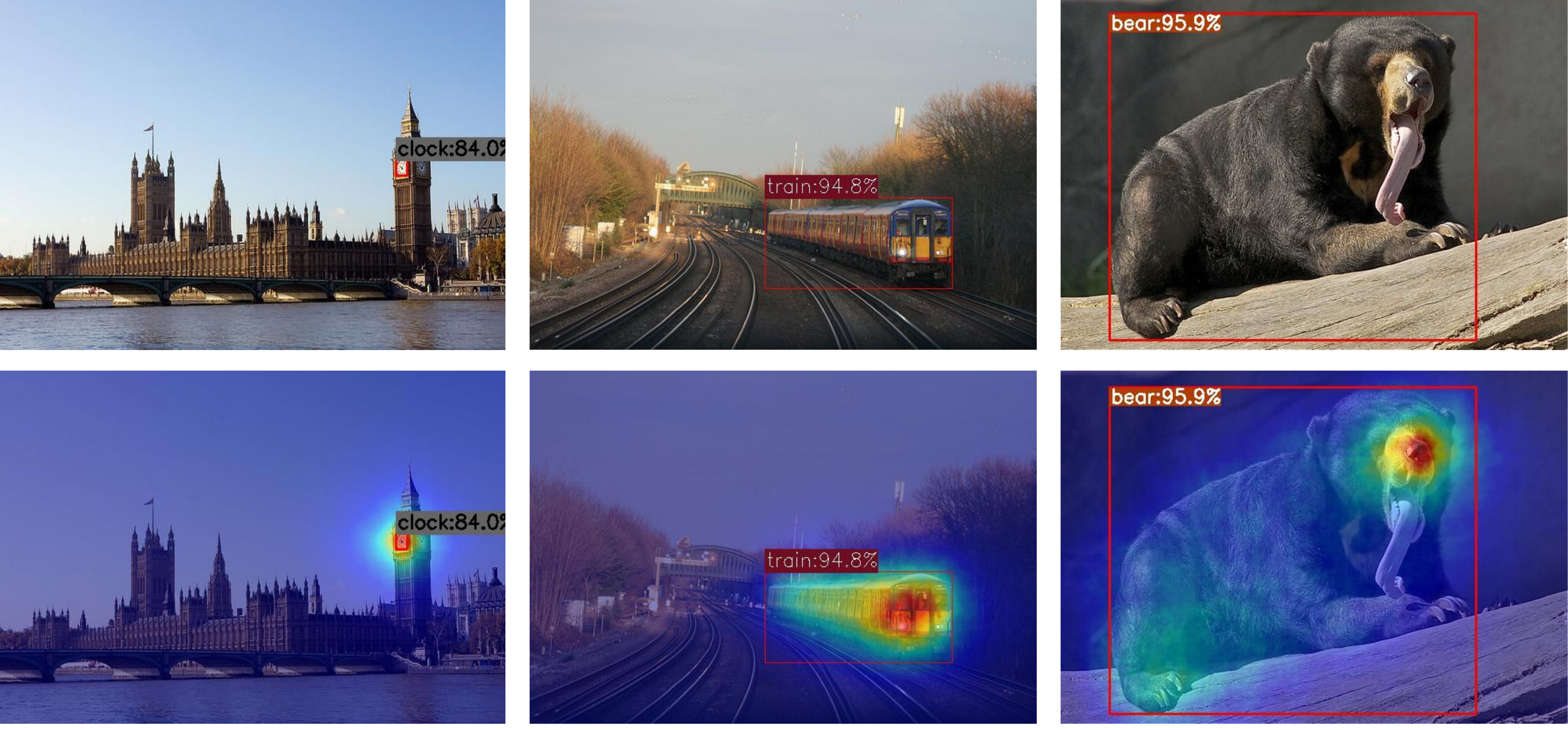}
    \caption{Saliency maps generated by the D-CLOSE method.}
    \label{fig:example}
\end{figure}

\begin{table}[hbt!]
\centering
\resizebox{.6\linewidth}{!}{%
    \begin{tabular}{ccc}
    \toprule
    \textbf{Method}          & \textbf{D-RISE} & \textbf{D-CLOSE} \\
    \midrule
    Running time (s) $\downarrow$ & 98.67 & \textbf{70} \\
    \bottomrule
    \end{tabular}%
}
\caption{\label{tab:time}Comparative evaluation in terms of inference time (seconds, averaged for each object) on MS-COCO validation set. The better is in bold.}
\end{table}

\subsection{Ablation studies}\label{sec:ablation}
Our method proposes to combine three important components, including multi-scale superpixel segment, density normalization, and multi-scale feature fusion. We perform the experiments to investigate the contribution level of each part to the final explanation performance. The results are reported in Table~\ref{ablation}.

\textbf{Superpixel Segment.} To validate the effectiveness of the superpixel segment step, we compare the evaluation metrics results using the superpixel segment results (row 1 of Table~\ref{ablation}) and the D-RISE results (Table~\ref{tab:quantitative_results}.)
With this, we help increase 1.07\% EBPG and 10.38\% over-all score.

\textbf{Density map.} In the proposed D-CLOSE, we have introduced density map normalization, which can produce smoother and more object-focused saliency maps. From Table~\ref{ablation} (row 2), we can see that the density map can boost 3.01 sparsity, 2.93\% EBPG, and 0.06\% over-all score, compared to just adding superpixel segment step.

\textbf{Multi-scale Feature Fusion.} As shown in Table~\ref{ablation}, we get impressive performance (21.09 sparsity, 22.03\% EBPG, 0.79\% over-all score) using multi-scale feature fusion. This result is comprehensible since this step removes the most noise and keeps only the most important features. Experiments also reflect that this step has the most significant influence on the final explanation. Finally, the amalgamation of the three aforementioned steps yields a result that is markedly superior to the outcome of implementing each step in isolation.

\section{Results}\label{sec:results}
As mentioned in Sec.~\ref{sec:introduction}, D-RISE is only effective when fine-tuning hyperparameters for individual objects. While our method can be applied to different sizes of objects with only one set of parameters for the entire dataset. 
To validate this, we first use 5000 images from the MS-COCO validation dataset~\cite{lin2014microsoft} and calculate the ratio of the bounding box size with the input image. 
We then use these ratios as input to the $k$-means clustering algorithm~\cite{na2010research} to divide the objects into three groups corresponding to the objects belonging to small, middle, and large groups. 
We calculate quantitative metrics for each of these groups and for the entire data set to demonstrate the effectiveness of D-CLOSE.
Based on the results in Table~\ref{tab:quantitative_results}, we observe that D-CLOSE outperforms Grad-CAM and D-RISE on all quantitative metrics. 
To the best of our knowledge, our evaluation is the first to calculate quantitative metrics for distinct object groups.
Besides, in Table~\ref{tab:time}, we also summarize the average inference time results for D-RISE and the proposed D-CLOSE. 
Our method is 1.4 times faster than D-RISE and achieves much better performance.

\section{Conclusions}\label{sec:conclusions}
In this paper, we introduced D-CLOSE, a new XAI method that can explain the decisions of any object detector. Our method samples the input with multiple levels of segmentation to make the explanation more stable and smooth. We proposed using quantitative metrics for each object group to demonstrate that our approach performs better than previous state-of-the-art methods. We conducted in-depth analyzes of the model’s prediction errors and the implementation of D-CLOSE in real-world images.
In future work, we want to optimize the computation time of XAI by removing redundant data samples during masking to debug and create workflows that improve model performance. Our code is available at \url{https://github.com/Binh24399/D-CLOSE}.
{\small
\bibliographystyle{ieee_fullname}
\bibliography{ref}

\begin{thebibliography}{10}\itemsep=-1pt

\bibitem{achanta2010slic}
Radhakrishna Achanta, Appu Shaji, Kevin Smith, Aurelien Lucchi, Pascal Fua, and
  Sabine S{\"u}sstrunk.
\newblock Slic superpixels.
\newblock Technical report, 2010.

\bibitem{adebayo2018sanity}
Julius Adebayo, Justin Gilmer, Michael Muelly, Ian Goodfellow, Moritz Hardt,
  and Been Kim.
\newblock Sanity checks for saliency maps.
\newblock {\em Advances in neural information processing systems}, 31, 2018.

\bibitem{bach2015pixel}
Sebastian Bach, Alexander Binder, Gr{\'e}goire Montavon, Frederick Klauschen,
  Klaus-Robert M{\"u}ller, and Wojciech Samek.
\newblock On pixel-wise explanations for non-linear classifier decisions by
  layer-wise relevance propagation.
\newblock {\em PloS one}, 10(7):e0130140, 2015.

\bibitem{bayer2021comparison}
Jens Bayer, David M{\"u}nch, and Michael Arens.
\newblock A comparison of deep saliency map generators on multispectral data in
  object detection.
\newblock In {\em Counterterrorism, Crime Fighting, Forensics, and Surveillance
  Technologies V}, volume 11869, pages 61--74. SPIE, 2021.

\bibitem{chattopadhay2018grad}
Aditya Chattopadhay, Anirban Sarkar, Prantik Howlader, and Vineeth~N
  Balasubramanian.
\newblock Grad-cam++: Generalized gradient-based visual explanations for deep
  convolutional networks.
\newblock In {\em 2018 IEEE winter conference on applications of computer
  vision (WACV)}, pages 839--847. IEEE, 2018.

\bibitem{cooper2021believe}
Jessica Cooper, Ognjen Arandjelovi{\'c}, and David~J Harrison.
\newblock Believe the hipe: Hierarchical perturbation for fast, robust and
  model-agnostic explanations.
\newblock {\em arXiv preprint arXiv:2103.05108}, 2021.

\bibitem{das2020opportunities}
Arun Das and Paul Rad.
\newblock Opportunities and challenges in explainable artificial intelligence
  (xai): A survey.
\newblock {\em arXiv preprint arXiv:2006.11371}, 2020.

\bibitem{do2018real}
Truong-Dong Do, Minh-Thien Duong, Quoc-Vu Dang, and My-Ha Le.
\newblock Real-time self-driving car navigation using deep neural network.
\newblock In {\em 2018 4th International Conference on Green Technology and
  Sustainable Development (GTSD)}, pages 7--12. IEEE, 2018.

\bibitem{ge2021yolox}
Zheng Ge, Songtao Liu, Feng Wang, Zeming Li, and Jian Sun.
\newblock Yolox: Exceeding yolo series in 2021.
\newblock {\em arXiv preprint arXiv:2107.08430}, 2021.

\bibitem{gomez2022metrics}
Tristan Gomez, Thomas Fr{\'e}our, and Harold Mouch{\`e}re.
\newblock Metrics for saliency map evaluation of deep learning explanation
  methods.
\newblock In {\em Pattern Recognition and Artificial Intelligence: Third
  International Conference, ICPRAI 2022, Paris, France, June 1--3, 2022,
  Proceedings, Part I}, pages 84--95. Springer, 2022.

\bibitem{gudovskiy2018explain}
Denis Gudovskiy, Alec Hodgkinson, Takuya Yamaguchi, Yasunori Ishii, and Sotaro
  Tsukizawa.
\newblock Explain to fix: A framework to interpret and correct dnn object
  detector predictions.
\newblock {\em arXiv preprint arXiv:1811.08011}, 2018.

\bibitem{hartley2021swag}
Thomas Hartley, Kirill Sidorov, Christopher Willis, and David Marshall.
\newblock Swag: Superpixels weighted by average gradients for explanations of
  cnns.
\newblock In {\em Proceedings of the IEEE/CVF Winter Conference on Applications
  of Computer Vision}, pages 423--432, 2021.

\bibitem{hoiem2012diagnosing}
Derek Hoiem, Yodsawalai Chodpathumwan, and Qieyun Dai.
\newblock Diagnosing error in object detectors.
\newblock In {\em European conference on computer vision}, pages 340--353.
  Springer, 2012.

\bibitem{kapishnikov2019xrai}
Andrei Kapishnikov, Tolga Bolukbasi, Fernanda Vi{\'e}gas, and Michael Terry.
\newblock Xrai: Better attributions through regions.
\newblock In {\em Proceedings of the IEEE/CVF International Conference on
  Computer Vision}, pages 4948--4957, 2019.

\bibitem{lin2014microsoft}
Tsung-Yi Lin, Michael Maire, Serge Belongie, James Hays, Pietro Perona, Deva
  Ramanan, Piotr Doll{\'a}r, and C~Lawrence Zitnick.
\newblock Microsoft coco: Common objects in context.
\newblock In {\em European conference on computer vision}, pages 740--755.
  Springer, 2014.

\bibitem{liu2016ssd}
Wei Liu, Dragomir Anguelov, Dumitru Erhan, Christian Szegedy, Scott Reed,
  Cheng-Yang Fu, and Alexander~C Berg.
\newblock Ssd: Single shot multibox detector.
\newblock In {\em European conference on computer vision}, pages 21--37.
  Springer, 2016.

\bibitem{lundberg2017unified}
Scott~M Lundberg and Su-In Lee.
\newblock A unified approach to interpreting model predictions.
\newblock {\em Advances in neural information processing systems}, 30, 2017.

\bibitem{miotto2018deep}
Riccardo Miotto, Fei Wang, Shuang Wang, Xiaoqian Jiang, and Joel~T Dudley.
\newblock Deep learning for healthcare: review, opportunities and challenges.
\newblock {\em Briefings in bioinformatics}, 19(6):1236--1246, 2018.

\bibitem{na2010research}
Shi Na, Liu Xumin, and Guan Yong.
\newblock Research on k-means clustering algorithm: An improved k-means
  clustering algorithm.
\newblock In {\em 2010 Third International Symposium on intelligent information
  technology and security informatics}, pages 63--67. Ieee, 2010.

\bibitem{nguyen2021evaluation}
Hung Truong~Thanh Nguyen, Hung~Quoc Cao, Khang Vo~Thanh Nguyen, and Nguyen
  Dinh~Khoi Pham.
\newblock Evaluation of explainable artificial intelligence: Shap, lime, and
  cam.
\newblock In {\em Proceedings of the FPT AI Conference}, pages 1--6, 2021.

\bibitem{nguyen2022secam}
Phong~X Nguyen, Hung~Q Cao, Khang~VT Nguyen, Hung Nguyen, and Takehisa Yairi.
\newblock Secam: Tightly accelerate the image explanation via region-based
  segmentation.
\newblock {\em IEICE TRANSACTIONS on Information and Systems},
  105(8):1401--1417, 2022.

\bibitem{nguyen2023towards}
Truong Thanh~Hung Nguyen, Van~Binh Truong, Vo~Thanh~Khang Nguyen, Quoc~Hung
  Cao, and Quoc~Khanh Nguyen.
\newblock Towards trust of explainable ai in thyroid nodule diagnosis.
\newblock {\em arXiv preprint arXiv:2303.04731}, 2023.

\bibitem{otsu1979threshold}
Nobuyuki Otsu.
\newblock A threshold selection method from gray-level histograms.
\newblock {\em IEEE transactions on systems, man, and cybernetics},
  9(1):62--66, 1979.

\bibitem{petsiuk2018rise}
Vitali Petsiuk, Abir Das, and Kate Saenko.
\newblock Rise: Randomized input sampling for explanation of black-box models.
\newblock {\em arXiv preprint arXiv:1806.07421}, 2018.

\bibitem{petsiuk2021black}
Vitali Petsiuk, Rajiv Jain, Varun Manjunatha, Vlad~I Morariu, Ashutosh Mehra,
  Vicente Ordonez, and Kate Saenko.
\newblock Black-box explanation of object detectors via saliency maps.
\newblock In {\em Proceedings of the IEEE/CVF Conference on Computer Vision and
  Pattern Recognition}, pages 11443--11452, 2021.

\bibitem{redmon2016you}
Joseph Redmon, Santosh Divvala, Ross Girshick, and Ali Farhadi.
\newblock You only look once: Unified, real-time object detection.
\newblock In {\em Proceedings of the IEEE conference on computer vision and
  pattern recognition}, pages 779--788, 2016.

\bibitem{regulation2018general}
General Data~Protection Regulation.
\newblock General data protection regulation (gdpr).
\newblock {\em Intersoft Consulting, Accessed in October}, 24(1), 2018.

\bibitem{ren2015faster}
Shaoqing Ren, Kaiming He, Ross Girshick, and Jian Sun.
\newblock Faster r-cnn: Towards real-time object detection with region proposal
  networks.
\newblock {\em Advances in neural information processing systems}, 28, 2015.

\bibitem{ribeiro2016should}
Marco~Tulio Ribeiro, Sameer Singh, and Carlos Guestrin.
\newblock " why should i trust you?" explaining the predictions of any
  classifier.
\newblock In {\em Proceedings of the 22nd ACM SIGKDD international conference
  on knowledge discovery and data mining}, pages 1135--1144, 2016.

\bibitem{sejr2021surrogate}
Jonas~Herskind Sejr, Peter Schneider-Kamp, and Naeem Ayoub.
\newblock Surrogate object detection explainer (sodex) with yolov4 and lime.
\newblock {\em Machine Learning and Knowledge Extraction}, 3(3):662--671, 2021.

\bibitem{selvaraju2017grad}
Ramprasaath~R Selvaraju, Michael Cogswell, Abhishek Das, Ramakrishna Vedantam,
  Devi Parikh, and Dhruv Batra.
\newblock Grad-cam: Visual explanations from deep networks via gradient-based
  localization.
\newblock In {\em Proceedings of the IEEE international conference on computer
  vision}, pages 618--626, 2017.

\bibitem{shrikumar2017learning}
Avanti Shrikumar, Peyton Greenside, and Anshul Kundaje.
\newblock Learning important features through propagating activation
  differences.
\newblock In {\em International conference on machine learning}, pages
  3145--3153. PMLR, 2017.

\bibitem{simonyan2014visualising}
Karen Simonyan, Andrea Vedaldi, and Andrew Zisserman.
\newblock Visualising image classification models and saliency maps.
\newblock {\em Deep Inside Convolutional Networks}, 2014.

\bibitem{sudhakar2021ada}
Mahesh Sudhakar, Sam Sattarzadeh, Konstantinos~N Plataniotis, Jongseong Jang,
  Yeonjeong Jeong, and Hyunwoo Kim.
\newblock Ada-sise: adaptive semantic input sampling for efficient explanation
  of convolutional neural networks.
\newblock In {\em ICASSP 2021-2021 IEEE International Conference on Acoustics,
  Speech and Signal Processing (ICASSP)}, pages 1715--1719. IEEE, 2021.

\bibitem{takumi2017multispectral}
Karasawa Takumi, Kohei Watanabe, Qishen Ha, Antonio Tejero-De-Pablos, Yoshitaka
  Ushiku, and Tatsuya Harada.
\newblock Multispectral object detection for autonomous vehicles.
\newblock In {\em Proceedings of the on Thematic Workshops of ACM Multimedia
  2017}, pages 35--43, 2017.

\bibitem{tsunakawa2019contrastive}
Hideomi Tsunakawa, Yoshitaka Kameya, Hanju Lee, Yosuke Shinya, and Naoki
  Mitsumoto.
\newblock Contrastive relevance propagation for interpreting predictions by a
  single-shot object detector.
\newblock In {\em 2019 International Joint Conference on Neural Networks
  (IJCNN)}, pages 1--9. IEEE, 2019.

\bibitem{wagner2019interpretable}
Jorg Wagner, Jan~Mathias Kohler, Tobias Gindele, Leon Hetzel, Jakob~Thaddaus
  Wiedemer, and Sven Behnke.
\newblock Interpretable and fine-grained visual explanations for convolutional
  neural networks.
\newblock In {\em Proceedings of the IEEE/CVF Conference on Computer Vision and
  Pattern Recognition}, pages 9097--9107, 2019.

\bibitem{wang2020score}
Haofan Wang, Zifan Wang, Mengnan Du, Fan Yang, Zijian Zhang, Sirui Ding, Piotr
  Mardziel, and Xia Hu.
\newblock Score-cam: Score-weighted visual explanations for convolutional
  neural networks.
\newblock In {\em Proceedings of the IEEE/CVF conference on computer vision and
  pattern recognition workshops}, pages 24--25, 2020.

\bibitem{xia2021receptive}
Pengfei Xia, Hongjing Niu, Ziqiang Li, and Bin Li.
\newblock On the receptive field misalignment in cam-based visual explanations.
\newblock {\em Pattern Recognition Letters}, 152:275--282, 2021.

\bibitem{yang2021mfpp}
Qing Yang, Xia Zhu, Jong-Kae Fwu, Yun Ye, Ganmei You, and Yuan Zhu.
\newblock Mfpp: Morphological fragmental perturbation pyramid for black-box
  model explanations.
\newblock In {\em 2020 25th International Conference on Pattern Recognition
  (ICPR)}, pages 1376--1383. IEEE, 2021.

\bibitem{zhang2021group}
Qinglong Zhang, Lu Rao, and Yubin Yang.
\newblock Group-cam: group score-weighted visual explanations for deep
  convolutional networks.
\newblock {\em arXiv preprint arXiv:2103.13859}, 2021.

\bibitem{zhang2019widerperson}
Shifeng Zhang, Yiliang Xie, Jun Wan, Hansheng Xia, Stan~Z Li, and Guodong Guo.
\newblock Widerperson: A diverse dataset for dense pedestrian detection in the
  wild.
\newblock {\em IEEE Transactions on Multimedia}, 22(2):380--393, 2019.

\bibitem{zhou2016learning}
Bolei Zhou, Aditya Khosla, Agata Lapedriza, Aude Oliva, and Antonio Torralba.
\newblock Learning deep features for discriminative localization.
\newblock In {\em Proceedings of the IEEE conference on computer vision and
  pattern recognition}, pages 2921--2929, 2016.

\bibitem{zou2019object}
Zhengxia Zou, Zhenwei Shi, Yuhong Guo, and Jieping Ye.
\newblock Object detection in 20 years: A survey.
\newblock {\em arXiv preprint arXiv:1905.05055}, 2019.

\end{thebibliography}
}
\end{document}